# Integrated Longitudinal and Lateral Control for High-Speed Platooning of Articulated Vehicles

Aashish Shaju, Steve Southward, and Mehdi Ahmadian

Center for Vehicle Systems and Safety (CVeSS), Virginia Tech, Blacksburg, VA 24060, USA

**Keywords:** Longitudinal Control, Clothoid-based path tracking, Adaptive lookahead distance, High-speed lateral control, Articulated vehicle control, Vehicle platooning.

## Abstract

This paper presents an integrated control framework for high-speed platooning of articulated vehicles, combining advanced longitudinal and lateral control strategies to address the challenges of safe and efficient autonomous driving. A novel proof of string stability for the proposed longitudinal control algorithm, based on a PID framework, ensures robust inter-vehicle distance regulation even under dynamic operating conditions. Additionally, an arc-length-based longitudinal error formulation enhances the applicability of the control system for curved roads, making it realistic for diverse driving scenarios. For lateral control, significant improvements to the clothoid-based path tracking algorithm are introduced to mitigate oscillations and enhance stability at high speeds. These enhancements include a tangential criterion for oscillation reduction, trajectory segment classification for adaptive lookahead distance tuning, a dual-adaptive mechanism for regulating the rate of lookahead distance adjustments based on both cross-track error (CTE) and anticipated road curvature, and extensions for articulated vehicle dynamics. The proposed integrated control framework is validated through simulations, demonstrating its effectiveness in maintaining stability, minimizing tracking errors, and ensuring robust performance in high-speed and challenging platooning scenarios.

## 1. Introduction

The advancement of autonomous vehicle technologies has paved the way for significant innovations in transportation, with platooning emerging as a promising strategy to enhance road safety, improve traffic efficiency, and reduce fuel consumption [1–3]. Platooning involves closely coordinated autonomous vehicles traveling in a formation, necessitating precise longitudinal and lateral control to maintain stability, minimize tracking errors, and ensure safety across various operating conditions. While considerable progress has been made in this domain, the challenges associated with high-speed operations, especially for articulated vehicles, remain a critical focus for researchers. This paper focuses on addressing the challenges of high-speed lateral control for articulated vehicles and the integration of longitudinal and lateral control for effective platooning.

Longitudinal control in vehicular platooning has seen extensive research aimed at ensuring internal and string stability across various scenarios, often using centralized or distributed PID variants [4]. Extensions to address communication delays and vehicle heterogeneity via robust methods e.g., H∞-based designs [5] or Lyapunov-based delay handling [6] have improved resilience under uncertainty. Recent works have explored adaptive sliding mode [7] and data-driven predictive control [8] to enhance performance in mixed-fleet scenarios. Experimental studies such as [9] have validated spacing policies in practical platoon deployments.

While notable progress has occurred, most existing research predominantly evaluates performance on straight-line trajectories, overlooking the complexities introduced by curved roads. By leveraging insights from [10,11], this work addresses such gaps by proposing an arc-length-based inter-vehicle spacing metric, enabling accurate control in curved road scenarios.

Lateral control under high-speed conditions must contend with tire slip, actuator limits, and rapid curvature changes factors that become more severe in articulated vehicles. Traditional path tracking controllers, such as pure pursuit or Stanley-based methods [12,13], often struggle with lookahead tuning, oscillation control, and poor curvature tracking. Model Predictive Control (MPC) has emerged as a strong candidate for lateral guidance [14], with applications extending to joint longitudinal-lateral integration [15] and articulated vehicles [16]. However, the high computational cost of nonlinear MPC and reliance on accurate vehicle models limit scalability. Alternative robust methods like sliding mode control [17,18] and learning-based fuzzy or neural architectures [19–21] show promise but often lack interpretability or formal guarantees. Few controllers are tailored to articulated configurations. Works like [22] demonstrate tube-based MPC for trailer sway mitigation, but such approaches remain domain-specific and are not easily extensible to general platooning operations.

Building upon advancements in longitudinal and lateral control, researchers have increasingly focused on integrating these systems for enhanced autonomous vehicle performance, particularly in platooning scenarios. Early frameworks [23] laid foundational concepts for integrated control, later extended by hierarchical MPC architectures [24,25] and combined control topologies [26,27]. However, most of these works target either straight roads or low-speed conditions. When applied to articulated vehicles, critical phenomena like rear-unit off-tracking, multi-body curvature mismatch, and dynamic stability degradation remain insufficiently addressed.

This paper bridges these identified gaps by developing and integrating robust, computationally efficient controllers tailored for high-speed articulated vehicle platooning. The core contributions are:

1. A formal string stability proof for a nested PID-based longitudinal controller, derived from an augmented state-space model that incorporates vehicle dynamics, offering a novel theoretical guarantee for this controller class [28].
2. High-speed enhancements to clothoid-based lateral control, including a tangential-error criterion and a dual-adaptive lookahead scheme that mitigate path oscillations and enable stable tracking in articulated vehicles [29].
3. An arc-length-based longitudinal error metric that replaces coordinate-based spacing calculations, enabling accurate inter-vehicle distance control on curved roads and improving real-world applicability

By validating this integrated framework through high-fidelity simulations on complex trajectories, this work demonstrates a practical and effective solution to a challenging problem in autonomous freight. The remainder of this paper details the control design, stability analysis, and comprehensive simulation results.

# 2. Longitudinal Control and Stability Analysis

## 2.1 Controller Architecture and Spacing Policy

The longitudinal control system employs a nested PID architecture to regulate inter-vehicle spacing, building upon the design detailed in our previous work [28]. The controller is structured with an inner loop dedicated to relative velocity control and an outer loop responsible for maintaining inter-vehicle distance. This hierarchical architecture, as shown in Figure 1, allows for decoupled control objectives, effectively reducing cross-coupling effects while enhancing the system's robustness against external disturbances.

The inner loop, a PI controller, regulates the relative velocity between the lead vehicle and the following vehicle. It was designed using loop shaping techniques to achieve optimal tracking of a predetermined reference velocity. The transfer function for this velocity controller is given by:

$$C_{velocity} = \frac{15612\, s + 151}{s}$$

The outer loop, a PD controller, manages the inter-vehicle distance. It uses the output of the inner loop to generate the required velocity adjustment for maintaining the desired spacing. The transfer function for the distance controller in the continuous time domain is:

$$C_{distance}(s) = \frac{10.85\, s + 14.25}{s + 43.39}$$

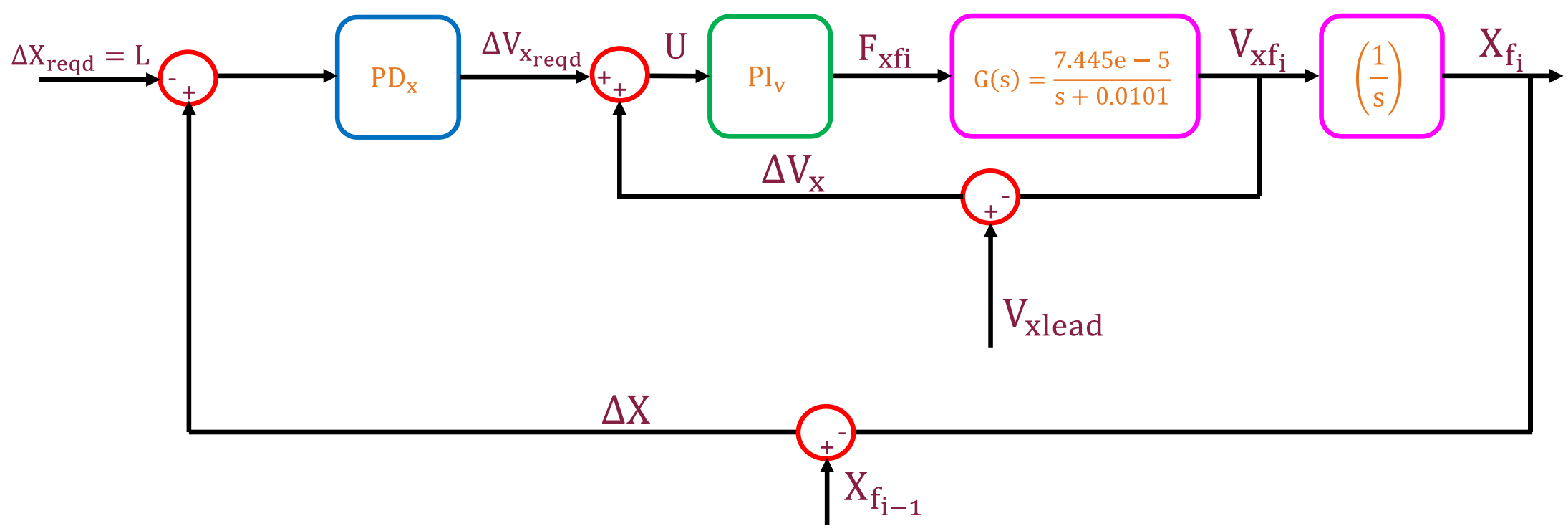

Figure 1 Nested PID controller structure for longitudinal control

The control framework operates under the constant time headway (CTH) spacing policy, a widely recognized approach for promoting string stability in vehicle platoons. The CTH policy dynamically adjusts the inter-vehicle distance based on the follower vehicle's velocity, thus providing enhanced stability and adaptability.

Consider a platoon with N+1 vehicles: one lead vehicle (i=0) and N follower vehicles (i=1,2…,N). The lead vehicle serves as a reference, with each follower vehicle adjusting its behavior based on information from both the lead and the preceding vehicle. The lead vehicle's position and velocity are denoted as $X_0(t)$ and $V_0(t)$ (interchangeably referred to as $V_{x_{lead}}$), while the position of follower vehicle i is $x_i(t)$. The goal is to control the position of each follower relative to its predecessor, maintaining a desired inter-vehicle spacing L, defined as:

$$L = d_{min} + t_{headway} * v_i \tag{1}$$

where $v_i$ is the ego vehicle velocity, $d_{min}$ is the minimum standstill distance (7.5 m) and $t_h$ is the time headway (0.3 s). The spacing error, $e_i$, for each follower vehicle i is then defined as the deviation from this desired spacing relative to the preceding vehicle i-1:

$$e_i(t) = x_{i-1}(t) - x_i(t) - L \tag{2}$$

Note that the platoon operates under a Vehicle-to-Vehicle (V2V) communication network, where each follower vehicle receives information from both the lead vehicle and its immediate predecessor. The lead vehicle broadcasts its position and velocity to all followers, while each follower communicates with its predecessor. This setup aligns with the Cooperative Adaptive Cruise Control (CACC) framework.

It is to be noted that only the longitudinal dynamics of platooning vehicles have been considered in this stability study. The longitudinal motion of the lead vehicle is characterized by:

$$\dot{X}_0(t) = V_0(t)$$

The leader can be driven by a human or can be autonomous, and it can act independent of the rest of the vehicles in the platoon.

## 2.2 State-Space Modeling for Platoon Dynamics

This section presents the derivation of a comprehensive state-space model representing the dynamics of an entire vehicle platoon. We begin by formulating the model for a single vehicle and its control system, then extending it to encompass multiple vehicles in a platoon formation. The primary goal is to construct a unified state-space representation that facilitates the analysis of internal and string stability. Internal stability will be assessed through eigenvalue analysis [4], while robustness of the dynamic model (and by extension, the controller) will be evaluated through basic robustness analysis, involving parametric variation in the eigenvalue analysis. The final and most rigorous component of this analysis involves proving string stability. We will employ the conventional definition, which ensures that the power (2-norm) of spacing error signals does not amplify downstream within the platoon.

### 2.2.1 Single-Vehicle Model Derivation

Consider a single vehicle with a nested control architecture comprising a velocity controller (inner loop) and a distance controller (outer loop) as shown in Figure 1. The longitudinal force in the Laplace domain is expressed as:

$$F_{xfi}(s) = \left(K_P + \frac{K_I}{s}\right) U(s)$$

$$\text{where } U(s) = \Delta V_{xreqd}(s) + V_{xlead}(s) - V_{xfi}(s)$$

Transforming to the time domain and introducing the integral state $x_1(t) = \int U(t)dt$, the velocity controller dynamics are given by:

$$\dot{x}_1(t) = U(t)$$

$$F_{xfi}(t) = K_P \dot{x}_1(t) + K_I x_1(t)$$

where:

- $F_{xfi}(t)$: Longitudinal force acting on the vehicle, which can be either positive or negative based on acceleration or braking.
- $x_1(t)$: State variable representing the integral of the velocity error.

Substituting $U(t)$ with its definition

$$U(t) = \Delta V_{xreqd}(t) + V_{xlead}(t) - V_{xfi}(t)$$

we obtain:

$$\dot{x}_1(t) = \Delta V_{xreqd}(t) + V_{xlead}(t) - V_{xfi}(t) \tag{3}$$

$$F_{xfi}(t) = K_I x_1(t) + K_P\left[\Delta V_{xreqd}(t) + V_{xlead}(t) - V_{xfi}(t)\right] \tag{4}$$

The longitudinal force $F_{xfi}(t)$ drives the dynamics of the vehicle. The relationship between the longitudinal force and the resulting velocity $V_{xfi}(t)$ is given by the transfer function:

$$\mathbf{G(s)} = \frac{\mathbf{k}}{\mathbf{s + c}}$$

Where k, and c are the system's gain and damping coefficients, respectively.

This translates into the following equation:

$$V_{xfi}(s) = \frac{k}{s+c} F_{xfi}(s)$$

Rewriting the transfer function in the time domain yields:

$$\dot{V}_{xfi}(t) + cV_{xfi}(t) = kF_{xfi}(t) \tag{5}$$

To include the velocity dynamics of the ego vehicle $V_{xfi}(t)$ into the state-space model, we define a new state variable:

$$x_2(t) = V_{xfi}(t)$$

From the vehicle dynamics equation 5, we can rewrite the velocity dynamics as:

$$\dot{x}_2(t) = -cx_2(t) + kF_{xfi}(t)$$

Substituting $F_{xfi}(t)$ from equation 4, we get,

$$\dot{x}_2(t) = -cx_2(t) + k\left(K_I x_1(t) + K_P\left[\Delta V_{xreqd}(t) + V_{xlead}(t) - V_{xfi}(t)\right]\right) \tag{6}$$

<u>Position Dynamics</u>

The position $X_{fi}(t)$ is related to velocity by:

$$\dot{X}_{fi}(t) = V_{xfi}(t)$$

Defining $x_3(t) = X_{fi}(t)$, we obtain:

$$\dot{x}_3(t) = V_{xfi}(t) = x_2(t) \tag{7}$$

<u>Distance Controller Dynamics</u>

The PD compensator's transfer function is given by:

$$\frac{y(s)}{u(s)} = \frac{K_{px}(s+z_x)}{s+p_x} = \frac{K_{px}s + K_{px}z_x}{s+p_x} \tag{8}$$

To represent this system in state-space form, we start with a general prototype:

$$\dot{x}(t) = [a]x(t) + [b]u(t)$$

$$y(t) = [c]x(t) + [d]u(t)$$

In the Laplace domain:

$$(s-a)x(s) = bu(s) \;\Rightarrow x(s) = \frac{b}{s-a} u(s)$$

Substituting $x(s)$ into the Laplace output equation we get:

$$y(s) = cx(s) + du(s) = \left[\frac{cb}{s-a} + d\right] u(s)$$

$$\Rightarrow \frac{y(s)}{u(s)} = \left[\frac{ds + (cb - da)}{s-a}\right]$$

Equating this result to the transfer function in equation 8.

$$\frac{y(s)}{u(s)} = \left[\frac{ds + (cb - da)}{s-a}\right] = \frac{K_{px}s + K_{px}z_x}{s+p_x}$$

Equating coefficients we get:

$$d = K_{px}$$

$$a = -p_x$$

$$cb - da = K_{px}z_x \Rightarrow cb + K_{px}p_x = K_{px}z_x \Rightarrow cb = K_{px}(z_x - p_x)$$

Since we have more unknowns than equations, we can arbitrarily choose the following:

$$b = 1 \Rightarrow c = K_{px}(z_X - p_x)$$

The final ss realization in terms of the original parameters is:

$$\dot{x}_4(t) = [-p_x]x_4(t) + u(t) \tag{9}$$

$$y(t) = K_{px}(z_x - p_x)x_4(t) + K_{px}u(t \tag{10}$$

where $y(t) = \Delta V_{x_{reqd}}$ and $u(t) = -L + X_{f_{i-1}} - X_{f_i}$

So, rewriting the above ss equations (9) and (10) in terms of the actual inputs and output variable we get:

$$\dot{x}_4(t) = [-p_x]x_4(t) - L + X_{f_{i-1}} - X_{f_i} = -p_x x_4 - L + X_{f_{i-1}} - x_3(t) \tag{11}$$

$$\Delta V_{x_{reqd}} = K_{px}(z_x - p_x)x_4(t) + K_{px}\left(-L + X_{f_{i-1}} - X_{f_i}\right) = K_{px}(z_x - p_x)x_4(t) + K_{px}\left(-L + X_{f_{i-1}} - x_3(t)\right) \tag{12}$$

Unified State-Space Model for a Single Vehicle

We substitute $\Delta V_{x_{reqd}} = K_{px}(z_x - p_x)x_4(t) + K_{px}\left(-L + X_{f_{i-1}} - x_3\right)$ from the PD controller into the state space equations previously derived in (3), (6), (7) and (11). The resulting state evolution equations are:

$$\dot{x}_1(t) = -x_2(t) - K_{px}x_3(t) + K_{px}(z_x - p_x)x_4(t) - K_{px}L + K_{px}X_{f_{i-1}} + V_{xlead} \tag{13}$$

$$\dot{x}_2(t) = kK_i x_1(t) - \left(c + kK_p\right)x_2(t) - kK_P K_{px}x_3(t) + kK_P K_{px}(z_x - p_x)x_4(t) - kK_P K_{px}L + kK_P K_{px}X_{f_{i-1}} + kK_P V_{xlead} \tag{14}$$

$$\dot{x}_3(t) = x_2(t) \tag{15}$$

$$\dot{x}_4(t) = -p_x x_4 - x_3(t) - L + X_{f_{i-1}} \tag{16}$$

We now assemble all the states and inputs into a temporary unified state-space model. Let:

State vector: $\mathbf{x_{temp}(t)} = [x_1(t),\ x_2(t),\ x_3(t),\ x_4(t)]^T$

Input vector: $\mathbf{u_{temp}(t)} = \left[L,\ X_{f_{i-1}}(t),\ V_{xlead}(t)\right]^T$

The state equation can be expressed as:

$$\dot{\mathbf{x}}_{\mathbf{temp}}(\mathbf{t}) = \begin{bmatrix} 0 & -1 & -K_{px} & K_{px}(z_x - p_x) \\ kK_I & -kK_P - c & -kK_P K_{px} & kK_P K_{px}(z_x - p_x) \\ 0 & 1 & 0 & 0 \\ 0 & 0 & -1 & -p_x \end{bmatrix} \mathbf{x_{temp}(t)} + \begin{bmatrix} -K_{px} & K_{px} & 1 \\ -kK_P K_{px} & kK_P K_{px} & kK_P \\ 0 & 0 & 0 \\ -1 & 1 & 0 \end{bmatrix} \mathbf{u_{temp}(t)}$$

Where,

$x_1$ – Integral state of the $PI_v$ velocity controller

$x_2$ – velocity of the ego vehicle $V_{xfi}$

$x_3$– Position of the ego vehicle $X_{f_i}$

$x_4$ – State corresponding to the PD compensator (Distance controller)

To improve the clarity and brevity of the state-space model while assembling them for the entire platoon, the variables were redefined and rearranged for better readability.

The new state variable is rearranged as:

$$\begin{bmatrix} x_i \\ v_i \\ \alpha_i \\ \beta_i \end{bmatrix} = \begin{bmatrix} x_3 \\ x_2 \\ x_1 \\ x_4 \end{bmatrix}$$

$x_i$ – Position of the ego vehicle $X_{f_i}$

$v_i$ – velocity of the ego vehicle $V_{xfi}$

$\alpha_i$–Integral state of the $PI_v$ velocity controller of the ego vehicle.

$\beta_i$ – State corresponding to the PD compensator (Distance controller) of the ego vehicle.

Also, $X_{f_{i-1}}$ is replaced by $X_{i-1}$, representing the position of the preceding vehicle.

$V_{xlead}$ is replaced by $V_0$, representing the velocity of the lead vehicle.

After this change of variables, the state evolution equations are expressed as follows:

1. Position Dynamics:

$$\dot{x}_i = v_i$$

2. Velocity Dynamics:

$$\dot{v}_i = \left(-K_P\, K_{px}\, k\right) x_i + \left(-c - K_P\, k\right) v_i + \left(K_I\, k\right) \alpha_i + \left(-K_P\, K_{px}\, k\, (p_x - z_x)\right)\, \beta_i - \left(K_P\, K_{px}\, k\right) L + \left(K_P\, k\right) V_0 + \left(K_P\, K_{px}\, k\right) X_{i-1}$$

3. PI Controller Dynamics:

$$\dot{\alpha}_i = \left(-K_{px}\right) x_i - v_i + \left(-K_{px}\,(p_x - z_x)\right)\, \beta_i - K_{px}\, L + V_0 + K_{px}\, X_{i-1}$$

4. PD Compensator Dynamics:

$$\dot{\beta}_i = -x_i + (-p_x)\, \beta_i - L + X_{i-1}$$

Now substituting for L from equation (1) (since we follow a constant time-headway-distance policy), the state space equation now becomes:

$$\begin{bmatrix} \dot{x}_i \\ \dot{v}_i \\ \dot{\alpha}_i \\ \dot{\beta}_i \end{bmatrix} = \begin{bmatrix} 0 & 1 & 0 & 0 \\ -kK_PK_{px} & -kK_PK_{px}t_{hd} - kK_P - c & kK_I & kK_PK_{px}(z_x - p_x) \\ -K_{px} & -K_{px}t_{hd} - 1 & 0 & K_{px}(z_x - p_x) \\ -1 & -t_{hd} & 0 & -p_x \end{bmatrix} \begin{bmatrix} x_i \\ v_i \\ \alpha_i \\ \beta_i \end{bmatrix} + \begin{bmatrix} 0 & 0 & 0 \\ -kK_PK_{px} & kK_PK_{px} & kK_P \\ -K_{px} & K_{px} & 1 \\ -1 & 1 & 0 \end{bmatrix} \begin{bmatrix} d_{min} \\ X_{i-1} \\ V_0 \end{bmatrix}$$

These equations are collectively represented in the final state-space form for a single vehicle as:

$$\dot{\mathbf{x}}_{\mathbf{i}}(\mathbf{t}) = \mathbf{A_0}\mathbf{x_i}(\mathbf{t}) + \mathbf{B_0}\mathbf{u_i}(\mathbf{t})$$

Where, $\mathbf{x_i} = \begin{bmatrix} x_i \\ v_i \\ \alpha_i \\ \beta_i \end{bmatrix}$, and $\mathbf{u_i} = \begin{bmatrix} d_{min} \\ X_{i-1} \\ V_0 \end{bmatrix}$

The matrices $\mathbf{A_0}$ and $\mathbf{B_0}$ are defined as:

$$\mathbf{A_0} = \begin{bmatrix} 0 & 1 & 0 & 0 \\ -kK_PK_{px} & -kK_PK_{px}t_{hd} - kK_P - c & kK_I & kK_PK_{px}(z_x - p_x) \\ -K_{px} & -K_{px}t_{hd} - 1 & 0 & K_{px}(z_x - p_x) \\ -1 & -t_{hd} & 0 & -p_x \end{bmatrix}$$

$$\mathbf{B_0} = \begin{bmatrix} 0 & 0 & 0 \\ -kK_PK_{px} & kK_PK_{px} & kK_P \\ -K_{px} & K_{px} & 1 \\ -1 & 1 & 0 \end{bmatrix}$$

### 2.2.2 Augmented Platoon Model

To construct the augmented model for a platoon of n+1 (n followers + 1 Lead) vehicles, the individual state-space models of each vehicle are coupled through the presence of the term $X_{i-1}$(position of the preceding vehicle) in the input vector of each individual vehicle state space equations. The individual vehicle state space model is first extended to a two-vehicle system, which serves as a basis for understanding the coupling between adjacent vehicles. Generalizing this approach to an n+1 vehicle platoon, the system's state-space representation integrates both the individual vehicle dynamics and the inter-vehicle coupling effects.

To ensure input independence, the position $X_0$ and velocity $V_0$ of the lead vehicle are restructured by introducing $X_0$ as a state variable ($x_u$) and defining its derivative as $V_0$. This modification ensures proper modeling and control design by eliminating redundancy in the input vector. The augmented state matrix $A_{aug}$ and input matrix $B_{aug}$ are updated to incorporate the new state variable and coupling dynamics between vehicles.

The final augmented state-space model (given below) encapsulates the dynamics of the entire platoon, including individual vehicle behavior, inter-vehicle coupling, and the influence of the lead vehicle.

$$[\dot{\mathbf{X}}] = [\mathbf{A_{aug}}][\mathbf{X}] + [\mathbf{B_{aug}}][\mathbf{u}]$$

Where:

$$\mathbf{X} = \begin{bmatrix} \mathbf{x_1} \\ \mathbf{x_2} \\ \vdots \\ \mathbf{x_n} \\ x_u \end{bmatrix}, \quad \mathbf{x_i} = \begin{bmatrix} x_i \\ v_i \\ \alpha_i \\ \beta_i \end{bmatrix}, \quad \mathbf{u} = \begin{bmatrix} d_{min} \\ V_0 \end{bmatrix}$$

1. **State Vector (X)**: Combines the states $\mathbf{x_i}$ for each follower vehicle (i=1,2,...,n) and the additional state variable $x_u$ representing the position ($X_0$) of the lead vehicle.
2. **Input Vector (u)**: Includes $d_{min}$ (desired minimum inter-vehicle spacing) and $V_0$ (lead vehicle velocity).
3. **Augmented State Matrix ($\mathbf{A_{aug}}$)**:

$$\mathbf{A_{aug}} = \begin{bmatrix} \mathbf{A_0} & \mathbf{0} & \mathbf{0} & \dots & \mathbf{0} & \mathbf{B_0}(:,\mathbf{2}) \\ \mathbf{A_{coup}} & \mathbf{A_0} & \mathbf{0} & \ddots & \mathbf{0} & \mathbf{0} \\ \mathbf{0} & \mathbf{A_{coup}} & \mathbf{A_0} & \ddots & \vdots & \vdots \\ \vdots & \ddots & \ddots & \ddots & \mathbf{0} & \vdots \\ \mathbf{0} & \dots & \mathbf{0} & \mathbf{A_{coup}} & \mathbf{A_0} & \vdots \\ \mathbf{0} & \mathbf{0} & \cdots & \cdots & \cdots & \mathbf{0} \end{bmatrix}_{(\mathbf{4n+1})*(\mathbf{4n+1})}$$

- $\mathbf{A_0}$: Captures the individual vehicle dynamics.
- $\mathbf{A_{coup}} = \begin{bmatrix} 0 & 0 & 0 & 0 \\ kK_PK_{px} & 0 & 0 & 0 \\ K_{px} & 0 & 0 & 0 \\ 1 & 0 & 0 & 0 \end{bmatrix}$ Represents coupling dynamics between adjacent vehicles due to platooning, ensuring the influence of a preceding vehicle on its follower.
- $\mathbf{B_0}(:,\mathbf{2})$: Accounts for the influence of the lead vehicle's position ($x_u$) on the dynamics.

4. **Augmented Input Matrix ($\mathbf{B_{aug}}$)**:

$$\mathbf{B_{aug}} = \begin{bmatrix} \mathbf{B_0}(:,[\mathbf{1},\mathbf{3}]) \\ \mathbf{B_{coup}}(:,[\mathbf{1},\mathbf{3}]) \\ \vdots \\ \mathbf{B_{coup}}(:,[\mathbf{1},\mathbf{3}]) \\ \mathbf{0} \quad \mathbf{1} \end{bmatrix}_{(\mathbf{4n+1})*\mathbf{2}}$$

- $\mathbf{B_0}(:,[\mathbf{1},\mathbf{3}])$: Represents the direct influence of the inputs ($d_{min}$, and $V_0$) on individual vehicle dynamics.
- $\mathbf{B_{coup}} = \begin{bmatrix} 0 & 0 & 0 \\ -kK_PK_{px} & 0 & kK_P \\ -K_{px} & 0 & 1 \\ -1 & 0 & 0 \end{bmatrix}$: Captures the input coupling effects between vehicles.

This model provides a foundation for subsequent stability analysis, offering insights into the platoon's overall behavior and performance under various conditions.

Now to formulate the output equation for spacing errors in a platoon, we define the spacing error for each vehicle as the difference between its position and that of the preceding vehicle, minus the desired inter-vehicle distance using equations 1 and 2. Using this definition, the individual error equations are systematically augmented to form a comprehensive output equation for the entire platoon, as detailed below.

$$\mathbf{y} = \mathbf{C_{aug}X} + \mathbf{D_{aug}u}$$

Where:

$$\mathbf{y} = \begin{bmatrix} e_1 \\ e_2 \\ \vdots \\ e_n \end{bmatrix}, \quad C_{aug} = \begin{bmatrix} \mathbf{C_0} & \mathbf{0} & \mathbf{0} & \dots & \dots & \mathbf{0} \\ \mathbf{C_{coup}} & \mathbf{C_0} & \mathbf{0} & \mathbf{0} & \dots & \mathbf{0} \\ \vdots & \mathbf{C_{coup}} & \mathbf{C_0} & \ddots & \ddots & \vdots \\ \vdots & \dots & \ddots & \ddots & \ddots & \mathbf{0} \\ \mathbf{0} & \mathbf{0} & \dots & \mathbf{C_{coup}} & \mathbf{C_0} & \mathbf{0} \\ \mathbf{0} & \mathbf{0} & \cdots & \cdots & \mathbf{C_{coup}} & \mathbf{C_0} \end{bmatrix},$$

$$\mathbf{D_{aug}} = \begin{bmatrix} \mathbf{D} \\ \mathbf{D} \\ \vdots \\ \vdots \\ \mathbf{D} \end{bmatrix}, \mathbf{C_0} = [-1 \quad -t_{hd} \quad 0 \quad 0], \mathbf{C_{coup}} = [1 \quad 0 \quad 0 \quad 0], \quad \text{and } \mathbf{D} = [-1 \quad 0]$$

The fidelity of this augmented model was confirmed by validating its output against a high-fidelity TruckSim® simulation, showing excellent agreement in vehicle velocities and inter-vehicle distances as shown in Figure 2.

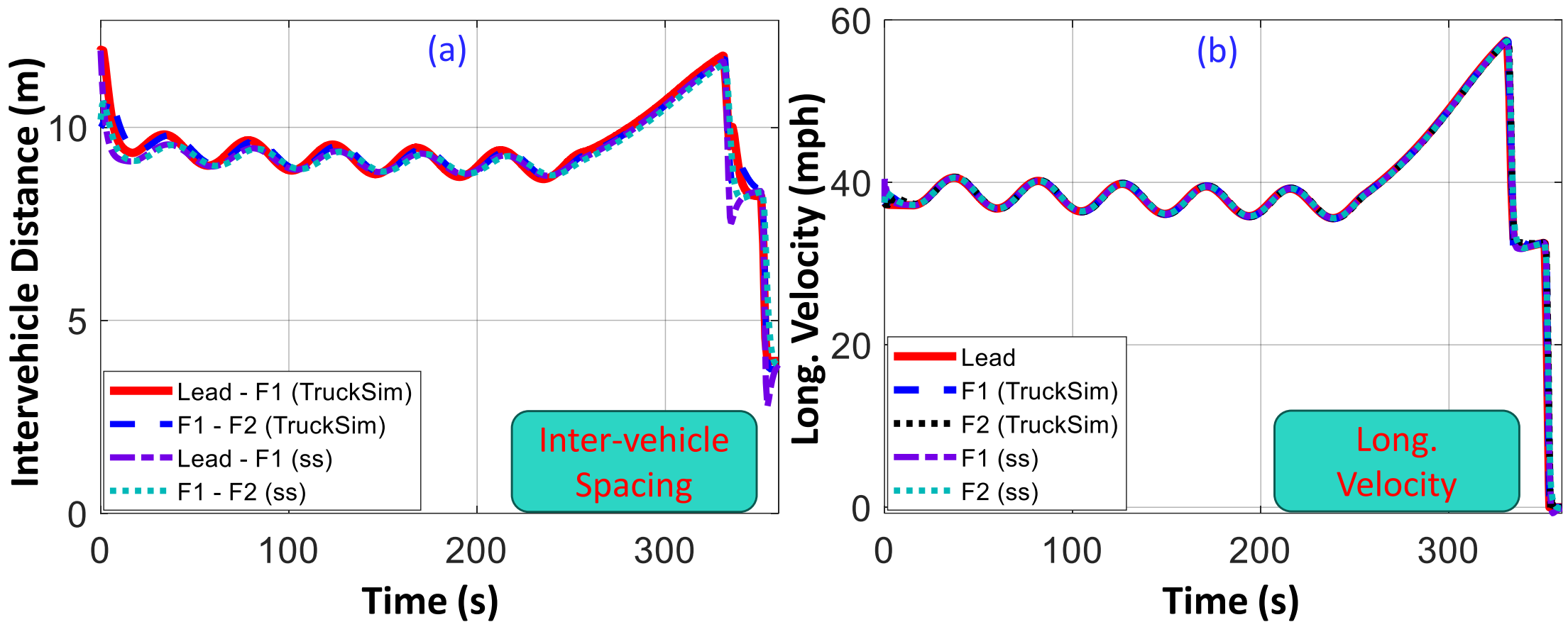

Figure 2 Comparison of (a) inter-vehicle spacing and (b) longitudinal velocity between the TruckSim®-Simulink® simulation and the augmented state-space model

## 2.3 Stability verification

### 2.3.3 Internal Stability (Eigen Value Analysis)

Eigenvalue analysis confirms internal stability for both homogeneous (Figure 3a) and heterogeneous (Figure 3b) platoons with up to 50 vehicles. The results depicted demonstrate that as the platoon size increases, the eigenvalues expand outward while remaining within stable regions, indicating stability for very large platoon configurations.

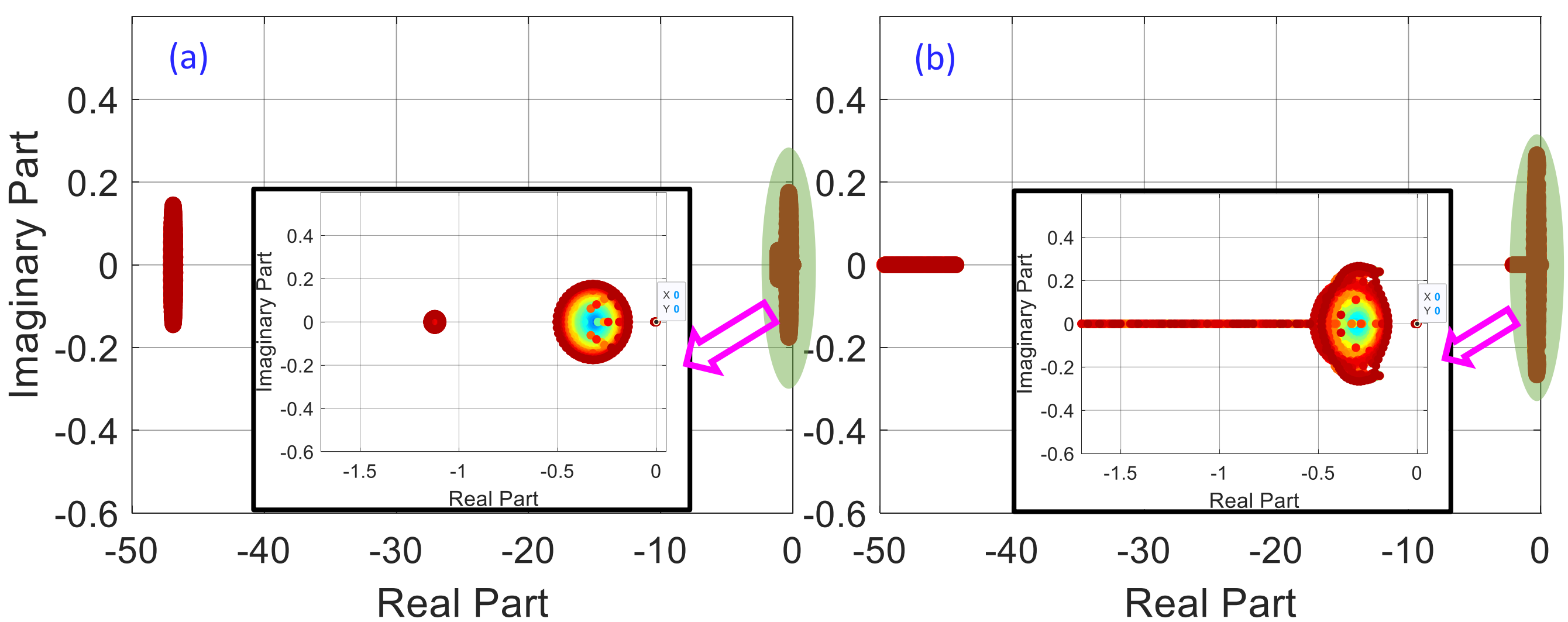

Figure 3 Eigenvalue distribution for the platoon system: zoomed-in views highlight the poles closest to the imaginary axis for both (a) homogeneous and (b) heterogeneous platoons

Both platoon types exhibit a zero eigenvalue at the origin, corresponding to the lead vehicle's position integration. This marginally stable mode does not compromise overall system stability. The remaining eigenvalues are located in the Left Half Plane (LHP), confirming asymptotic stability. Heterogeneous platoons maintain stability despite parameter variations, as evidenced by eigenvalues remaining in stable regions. To evaluate robustness, key vehicle parameters were varied by 50%. These perturbations, incorporated into the state-space matrices $A_{aug}$ and $B_{aug}$, resulted in minor eigenvalue shifts without causing instability. The absence of eigenvalue migration to the Right Half Plane (RHP) confirms the system's internal stability and the controller's robustness to parameter uncertainties.

### 2.3.4 String Stability (Frequency Response Analysis)

String stability ensures that disturbances in spacing and velocity errors do not amplify as they propagate downstream in a platoon. Mathematically, this requires the ratio of consecutive error transfer functions, $F(s) = \frac{Error_i(s)}{Error_{i-1}(s)}$, to satisfy $||F(j\omega)||_2 \leq 1$. This condition prevents error growth across the platoon.

Using the augmented state-space model, the Laplace transform gives the input-output transfer function:

$$\mathbf{y(s) = E(s) = H(s)u(s);\ where\ H(s) = \left[C_{aug}[sI - A_{aug}]^{-1}B_{aug} + D_{aug}\right]}$$

For a three-vehicle platoon (one leader and two followers), the error transfer functions are:

$$E(s) = \begin{bmatrix} e_1(s) \\ e_2(s) \end{bmatrix} = \begin{bmatrix} H_{11}(s) & H_{12}(s) \\ H_{21}(s) & H_{22}(s) \end{bmatrix} \begin{bmatrix} d_{min}(s) \\ V_0(s) \end{bmatrix}$$

This results in:

$$e_1(s) = H_{11}(s)d_{min}(s) + H_{12}(s)V_0(s)$$

$$e_2(s) = H_{21}(s)d_{min}(s) + H_{22}(s)V_0(s)$$

Focusing on velocity fluctuations $V_0(s)$, the ratios $\frac{e_1(s)}{V_0(s)}$ and $\frac{e_2(s)}{V_0(s)}$. remain below 0 dB (Figure 4a), confirming no amplification of lead vehicle disturbances.

String stability is confirmed if spacing errors attenuate downstream, requiring $\left\|\frac{e_i(s)}{e_{i-1}(s)}\right\|_2 \leq 1$. For a three-vehicle platoon:

$$\frac{e_2(s)}{e_1(s)} = \frac{H_{21}(s)D_{min} + H_{22}(s)v0}{H_{11}(s)D_{min} + H_{12}(s)v0}$$

Assuming step inputs for $d_{min}(s) = \frac{D_{min}}{s}$ and $V_0(s) = \frac{v0}{s}$ the ratio confirms the string stability condition (Figure 4b). This demonstrates that disturbances decay downstream, ensuring string stability for the platoon.

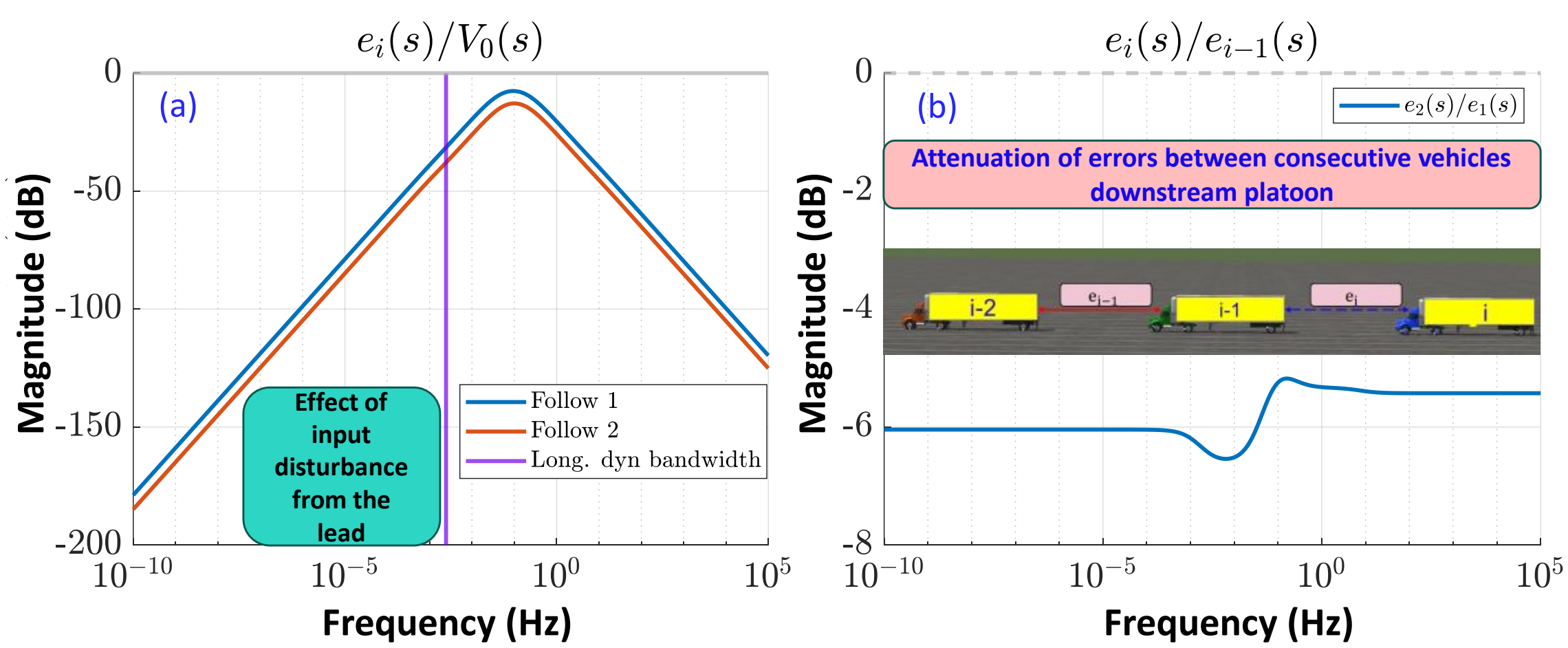

Figure 4 Frequency response for string stability. (a) Magnitude of $\frac{e_i(s)}{V_0(s)}$ for the first and second followers, showing no amplification of lead vehicle disturbances. (b) Magnitude of $\frac{e_2(s)}{e_1(s)}$, confirming decay of spacing errors downstream

## 2.4 Arc-Length Adaptation for Curved Roads

A key contribution of this work is the adaptation of the longitudinal controller for accurate operation on curved roads. Traditional error calculations based on a single coordinate axis are insufficient for such paths, as the true distance traveled along the road differs from the straight-line distance between vehicles. To address this, an arc-length-based error formulation was implemented to regulate the true on-road distance.

The method works by having each follower vehicle dynamically construct a virtual path representing the recent trajectory of its preceding vehicle. This is achieved by continuously recording and storing a moving window of the preceding vehicle's rear axle position coordinates, received via V2V communication. At each time step, the longitudinal error is calculated not as a simple difference in coordinates, but by numerically integrating the distance along this reconstructed virtual path. This integration is performed from the ego vehicle's current position to the preceding vehicle's current position, ensuring the controller regulates the true traveled distance.

This simple yet effective modification enhances the controller's applicability to realistic driving environments without requiring a pre-defined route map or altering the core PID control law, thus eliminating a major limitation of many CACC systems. Unlike Frenet-frame formulations that measure longitudinal spacing along a predefined road centerline and decouple motion into (s, d) coordinates, the proposed arc-length-based metric reconstructs the leader's actual trajectory in real time using V2V data. This eliminates the need for explicit road geometry, enabling accurate spacing control on curved paths even in map-free platooning scenarios.

# 3. High Speed Lateral Control for Articulated Vehicles

The lateral control system in this study is founded on a clothoid-based path tracking controller from our previous work [29]. Requiring only the reference path coordinates and the vehicle's current pose, the algorithm generates an optimal clothoid trajectory that aligns with the path by combining principles from Pure Pursuit and Stanley controllers. Clothoids were selected over other geometric primitives due to their inherent $C^2$ continuity, which provides a continuously changing curvature essential for generating smooth steering commands and maintaining stability at high speeds. The controller employs local path planning, iteratively searching through a predefined range of lookahead distances to find the optimal clothoid that minimizes lateral deviation while ensuring smooth transitions. The lookahead distance, which represents the target point on the desired trajectory, is a critical parameter that balances control responsiveness with smoothness.

While effective at speeds up to 25 mph, the original controller showed oscillations and overshoot in high-speed curved-to-straight transitions due to slow adaptation to faster dynamics, motivating the need for high-speed enhancements.

To address the limitations of the baseline controller, several key enhancements were developed and integrated.

**1. Integration of Tangential Criteria:** To mitigate the high-speed oscillations observed in scenarios like a dual-lane change, a tangential criterion was implemented alongside the standard Fréchet distance minimization. These oscillations often persisted despite small cross-track errors because the fitted clothoids would intersect the reference path at an angle rather than approaching it tangentially. The criterion calculates the angle between the reference path's slope and the clothoid's slope at the target point. If this angle exceeds a 5-degree threshold, the lookahead distance is iteratively adjusted to find a more tangential solution. This approach significantly reduces trajectory overshoot and improves settling time during high-speed maneuvers (Figure 5a).

**2. Trajectory Segment Classification:** To make the controller context-aware, a real-time path segmentation and classification (Figure 5b) mechanism was developed. Since direct curvature calculation from raw waypoint data can be noisy, the reference trajectory is first resampled at evenly spaced arc-length intervals. A quadratic polynomial is then fitted to these points to compute a smooth and reliable curvature profile. Based on the mean curvature of a 14-meter segment ahead of the vehicle, the path is classified in real-time as Straight (|mean curvature| < 0.01), Moderate Curve (0.01–0.03), or Tight Curve (≥ 0.03).

This classification allows for context-aware adaptation of the lookahead distance. The search range is dynamically adjusted based on the segment type: a longer range for straight paths (11–13 m) to reduce over-corrections, and a shorter range for tight curves (6–7 m) to ensure close tracking and minimize errors.

The selection of these specific numerical parameters is grounded in a combination of control-theoretic principles and empirical validation. The 14-meter segment length was chosen to slightly exceed the controller's maximum possible lookahead distance of 13 meters. This maximum lookahead was itself determined through extensive sensitivity analysis across a wide range of speeds and road geometries. It represents an optimal trade-off: increasing the lookahead beyond 13 m yielded no discernible improvement in path-following smoothness, even during high-speed straight-line driving, while unnecessarily expanding the computational search space. Conversely, reducing the maximum lookahead below this value induced high-frequency oscillations at highway speeds (~75 mph). Therefore, analyzing a 14-meter segment ensures the classification is based on the full extent of the path that the controller might consider for its lookahead target, making the classification directly relevant to the control action.

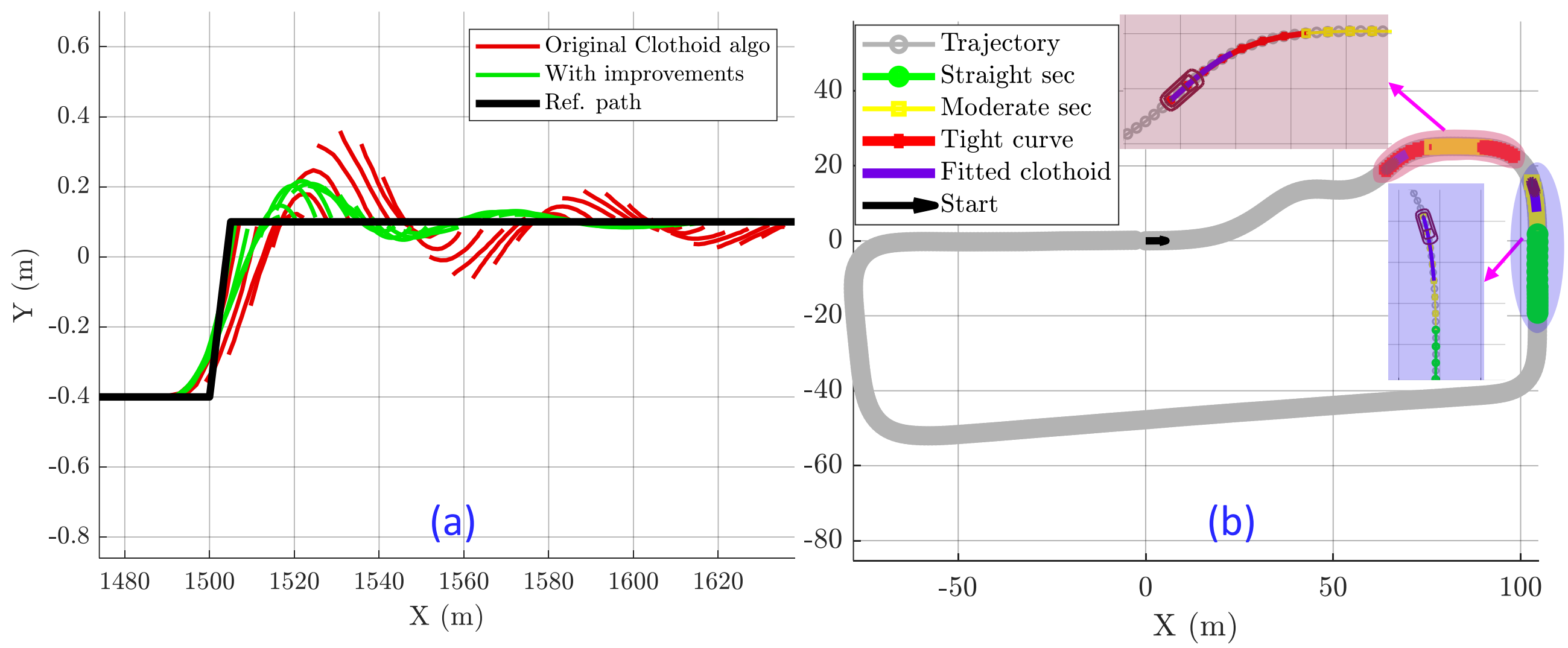

Figure 5 Illustration of the key improvements to the baseline clothoid controller (a) Oscillation damping of generated clothoids at each time steps through tangential check, (b) dynamic lookahead range based on segment classification

The curvature thresholds were established based on geometric principles and then validated empirically. A mean curvature below 0.01 $m^{-1}$ corresponds to a radius of curvature greater than 100 meters, which is geometrically characteristic of highway straights or very gentle curves. In these scenarios, a longer lookahead is safe and desirable for smoothness. Conversely, a curvature exceeding 0.03 $m^{-1}$ corresponds to a radius of curvature less than 33 meters, indicative of sharp turns or hairpin-style curves that demand a shorter, more reactive lookahead to prevent corner-cutting. The intermediate range captures typical highway curves. While these values were fine-tuned across numerous simulated tracks, they are fundamentally grounded in the geometric properties of distinct road types.

This classification (Figure 5b) allows for context-aware adaptation of the lookahead distance search range, which is dynamically adjusted to balance the competing objectives of path conformance and control smoothness:

- **Straight Paths (11–13 m):** A longer lookahead range is used here because it encourages the controller to generate clothoids with very low curvature. This results in smoother, less aggressive steering inputs, which is critical for enhancing stability and ride comfort during high-speed travel.

- **Moderate Curves (8–10 m):** A mid-range lookahead provides an essential balance, allowing the controller to follow the curve accurately without causing the abrupt steering actions that a very short lookahead would necessitate, nor the path deviation that a long lookahead might cause.
- **Tight Curves (6–7 m):** A shorter lookahead range is critical in this regime. It forces the controller to generate a more sharply curved clothoid that closely conforms to the reference path's high curvature. This prioritizes tracking accuracy and minimizes path deviation (i.e., corner-cutting), which is the primary control objective in sharp turns.

**3. Implementation of a Dual-Adaptive Lookahead Mechanism**: The aforementioned enhancements are unified within a dual-adaptive mechanism that dynamically adjusts the lookahead distance based on both cross-track error (CTE) and the anticipated path classification (Figure 6a and b). This approach ensures smooth transitions and robust tracking by adapting the control target in real-time to both anticipated road geometry and current vehicle dynamics.

The mechanism's logic operates through two concurrent strategies:

a. Reactive CTE-Based Adaptation: The controller simultaneously responds to the vehicle's real-time tracking performance. A large CTE prompts a reduction in the lookahead distance to prioritize rapid error correction and regain the reference path quickly. Conversely, as the CTE diminishes and the vehicle stabilizes, the lookahead distance is increased to promote smoother, less aggressive steering inputs, enhancing ride comfort (Figure 6a).
b. Proactive Segment-Based Adaptation: The path classification (Straight, Moderate, Tight) proactively sets the nominal upper ($R_{max}$) and lower ($R_{min}$) bounds for the lookahead distance search space. To ensure stability during transitions between different road types (e.g., entering a curve), these bounds are adjusted gradually with a controlled rate of change, preventing abrupt shifts in controller behavior (Figure 6b).

By combining these proactive and reactive strategies, the controller operates predictively based on upcoming road curvature while remaining responsive to immediate tracking errors. This dual approach is critical for maintaining stability and accuracy across the diverse and challenging conditions of high-speed driving.

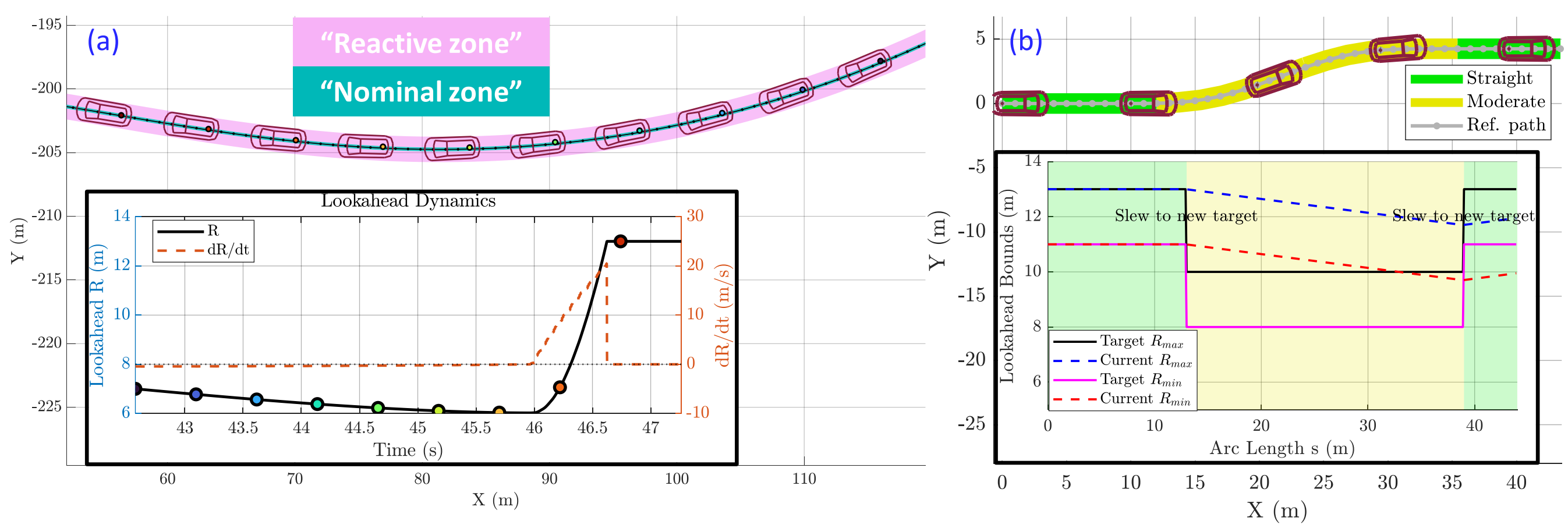

Figure 6 Dual-Adaptive Lookahead Dynamics. (a) Reactive Adaptation: From a real simulation, as tracking error is corrected (t=43-46s), the lookahead R is held low. Once on path (t>46s), R and its rate dR/dt increase to restore smoothness. (b) Proactive Adaptation: The controller's current bounds, $R_{max}$ and $R_{min}$, slew at a controlled rate towards the new target bounds (stepped lines) dictated by the upcoming path classification (shaded regions), ensuring smooth transitions.

**4. Application and Optimization for Articulated Vehicles:** While the enhanced clothoid-based controller was developed for single-unit vehicles, its extension to articulated systems—such as semi-tractor trailers—requires additional considerations. In articulated vehicles, the tractor and trailer follow distinct path curvatures, particularly in tight curves, leading to misalignment between segments. If unaddressed, this disparity can degrade path-following accuracy and stability.

To manage these dynamics, the controller was adapted to track alternative points on the tractor: steer axle, fifth wheel, and rear axle (tested on a day cab tractor, 300 HP, with a 45-ft trailer). The existing lead-filter-compensated framework [29] enabled seamless integration, requiring only a steady-state mapping between steering input and curvature at the chosen point.

This mapping was derived from extensive simulations covering a range of speeds and steering angles, producing lookup tables for each point. Robust filtering ensured accurate forward and inverse curvature mappings despite estimation noise, enabling precise clothoid generation.

A sensitivity analysis was conducted using:

- Dual Lane Change (DLC) - evaluating high-speed performance (Figure 7a).
- Figure-8 trajectory - assessing tight-curve handling (Figure 7b).

Cross-track error (CTE) was measured for points along the line between the steer axle (origin) and rear axle (3.5 m aft), as depicted in Figure 7c. Results (Figure 7d and e) showed the steer axle consistently achieved the lowest combined CTE, reducing oscillations and maintaining stability across both maneuvers. While the fifth wheel occasionally reduced tractor CTE, the steer axle provided superior overall performance.

Accordingly, all subsequent high-speed lateral control simulations for articulated vehicles were configured to track the steer axle, ensuring robust performance across the complex curvature dynamics of multi-body systems.

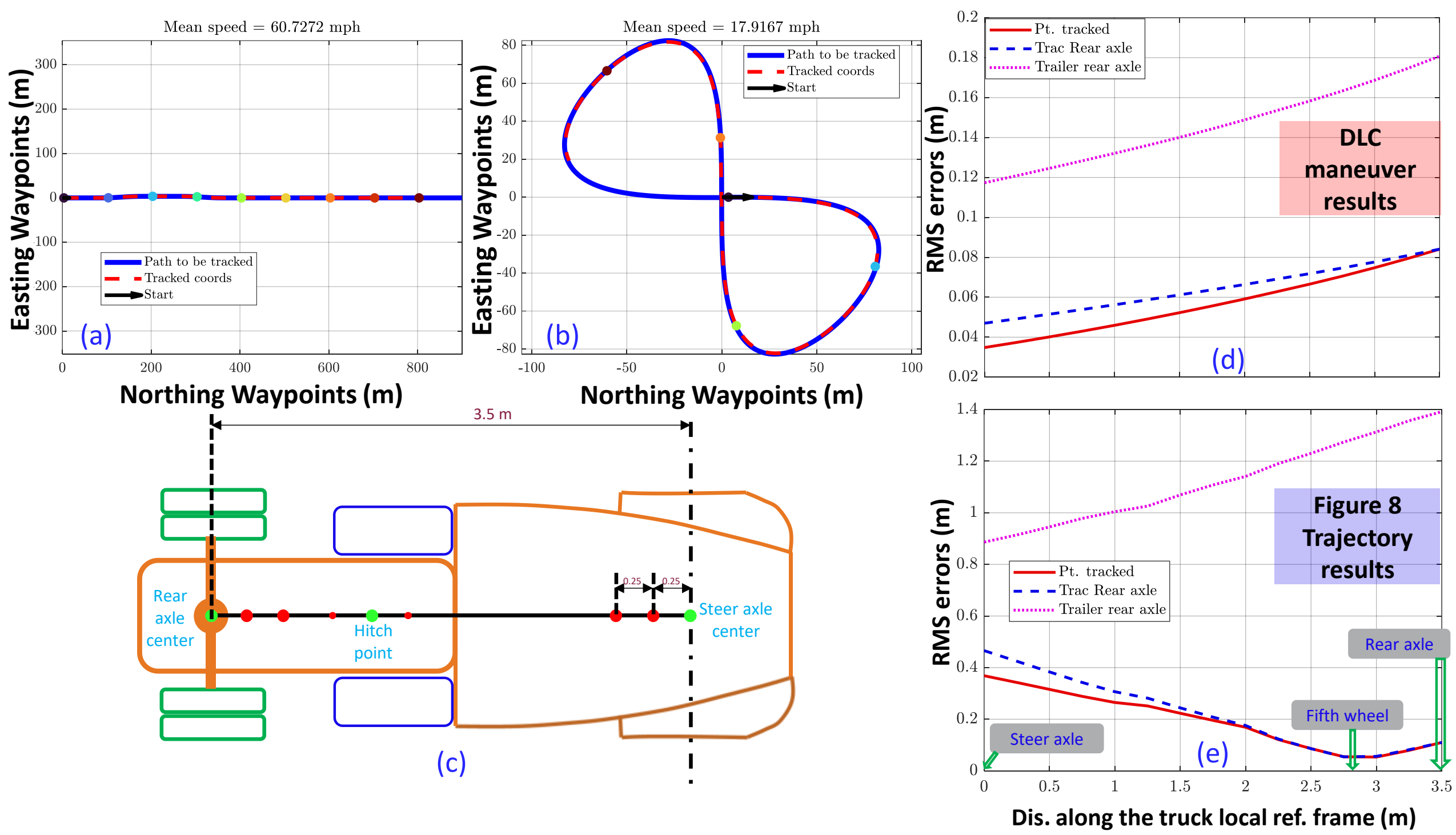

Figure 7 Application of the enhanced clothoid controller to articulated vehicles: (a, b) trajectories for the DLC and figure-8 maneuvers, (c) candidate tracking points along the tractor body (steer axle, fifth wheel/hitch point, rear axle), and (d, e) RMS cross-track error analysis across these points

## 5. Integrated Control Framework for Platooning

The complete integrated control framework, shown in Figure 8, seamlessly combines the longitudinal and lateral controllers to enable efficient and stable platooning of articulated vehicles. The system architecture is intentionally modular, maintaining distinct control loops for the longitudinal and lateral dynamics that operate on a shared reference trajectory. This allows each controller to function with its core logic unchanged. The key enabler for this integration on curved roads is the adaptation of the longitudinal controller to use the arc-length-based error calculation (as described in Section 2.4), replacing the simple coordinate-based difference. This modification allows the robust PID controller to manage inter-vehicle spacing accurately on complex geometries without requiring a complete redesign.

Within this framework, the enhanced lateral controller generates optimal clothoids for precise path tracking, while the longitudinal controller simultaneously regulates inter-vehicle spacing. This decoupled approach effectively achieves robust platooning capabilities with minimal development effort, demonstrating the system's adaptability for multi-vehicle coordination.

The performance of the integrated framework was validated in a high-fidelity co-simulation environment, using TruckSim® for the nonlinear vehicle dynamics and Simulink® for the controller implementation. The study utilized a three-vehicle platoon of identical semi-tractor trailers (300 HP tractors with 45-foot trailers) on a surface with a nominal friction coefficient of 0.85. The lead vehicle was tasked with tracking the reference trajectory and a predefined speed profile. Following vehicles received this reference path for lateral control, along with state information from the lead and preceding vehicles for longitudinal control. The controllers operated at a sampling frequency of 100 Hz, while the TruckSim® plant model updated at 1000 Hz.

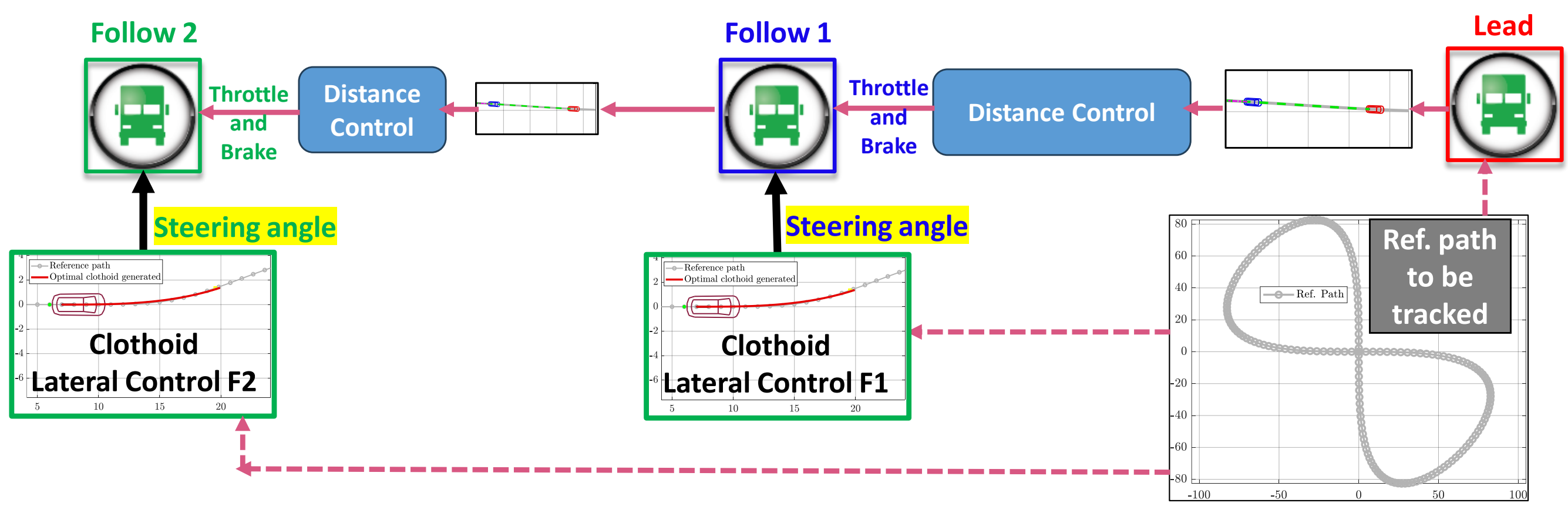

Figure 8 Integrated Control Framework: Combining both the lateral and longitudinal controllers for platooning

## 4. Simulation Results and Analysis

The performance of the integrated control framework was evaluated across three distinct and challenging trajectories designed to stress different aspects of the system:

1) a high-speed Dual Lane Change (DLC) maneuver to test stability during rapid lateral shifts.

2) a moderate-speed traversal of the Blue Ridge Parkway to evaluate performance on sustained curves.

3) low-speed navigation of the South Grade Road, featuring tight hairpin turns that test handling at the limits of vehicle articulation.

In each scenario, we analyze key longitudinal metrics (spacing error, speed, control inputs) and lateral metrics (cross-track error at key vehicle points, steering angle, lateral acceleration). It is to be noted that the colored markers (●, ●,…), spaced at 200-meter intervals, serve as spatial references throughout each simulation. These markers are instrumental in correlating spatial data with time-domain results such as spacing errors, cross-track errors (CTEs), and lateral accelerations, etc.

## 4.1 Dual Lane Change Maneuver

The dual lane change (DLC) maneuver serves as a critical test for assessing platoon stability and coordination during a rapid, high-speed lateral transition. The longitudinal control performance, illustrated in Figure 9, highlights the system's behavior under maximum acceleration. As shown in the platoon trajectory Figure 9a), the vehicles maintain a cohesive formation. However, the spacing error plot (Figure 9b) reveals a key insight: while Follower 1's error remains largely within the desired ±2 m bounds, the error for Follower 2 grows beyond this limit. This is an expected outcome of the test's extreme design. The speed profiles (Figure 10b) confirm that all vehicles are accelerating at their maximum capacity. Because the vehicles are identical, Follower 2, despite also being at full throttle, has no additional acceleration capacity to close the gap created by the lead vehicle. This scenario was intentionally designed to evaluate the system's stability under saturation, demonstrating that even when pushed to its physical limits, the platoon remains stable without unsafe spacing reductions.

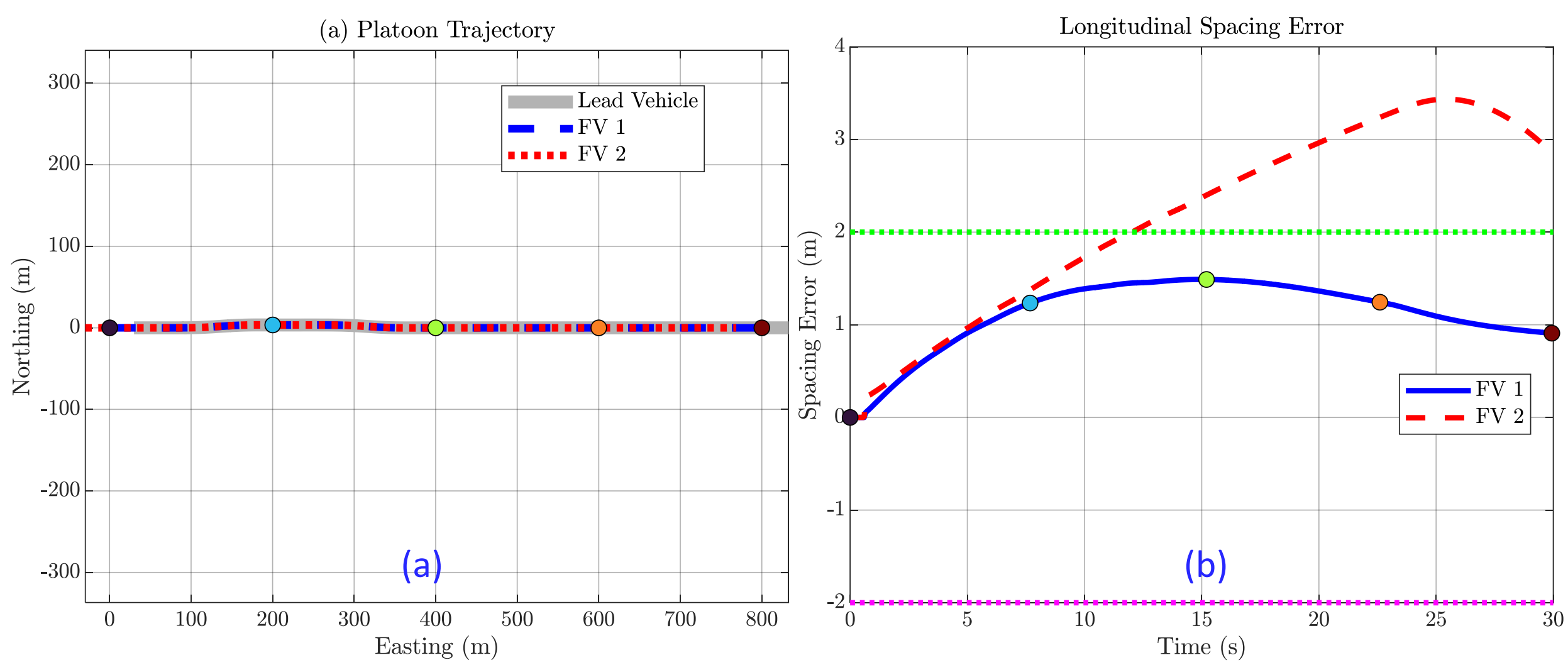

Figure 9 Longitudinal control results for DLC maneuver: (a) Trajectory of the lead and following vehicles, highlighting key spatial markers (b) Spacing error: Follow Vehicle 1 remains within ±2 m bounds, while Follow Vehicle 2 exceeds limits during acceleration

Lateral performance during this high-stress maneuver remains highly robust. The cross-track errors (CTEs) for all key points on both follower vehicles (Figure 10a) are tightly controlled, with maximum deviations staying below 0.35 m. Crucially, the error magnitudes for Follower 2 do not amplify compared to Follower 1, indicating strong lateral string stability—a key objective for platoon safety. Furthermore, the corresponding lateral accelerations (Figure 10b) remain well within safe comfort boundaries.

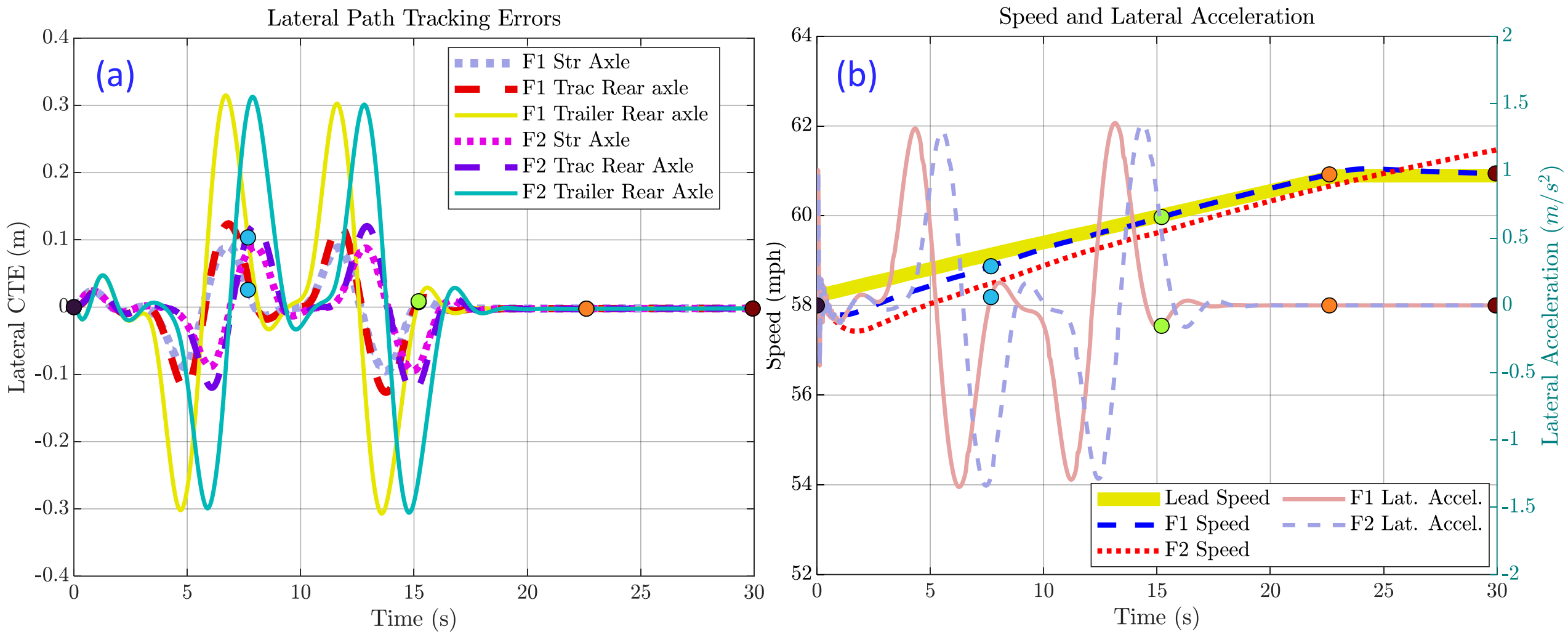

Figure 10 Lateral performance during the High-Speed DLC maneuver: (a) cross-track errors (CTEs) for tractor and trailer reference points on both follower vehicles remain within ±0.35 m, indicating strong lateral tracking (b) speed profiles (left axis) show coordinated acceleration across the platoon, while lateral accelerations (right axis) stay well within comfort limits

## 4.2 Blue Ridge Parkway Scenario (VA)

This scenario evaluates the controller's performance on a realistic highway trajectory characterized by moderate curves and straight segments at speeds up to the 45-mph limit. Longitudinal performance, depicted in Figure 11, demonstrates the system's robust response to dynamic speed changes. While the platoon maintains a cohesive formation along the winding path (Figure 11a), the spacing error for the follower vehicles (Figure 11b) shows transient positive spikes that occasionally exceed the +2 m bound. This is an expected behavior attributed to the “accordion effect” in curves: when the lead vehicle exits a turn and accelerates, the followers, still navigating the curve at a lower speed, cannot match its acceleration instantly. This creates a temporary gap that is quickly corrected once they exit the turn. Critically, the spacing error remains positive throughout, ensuring there is no risk of collision.

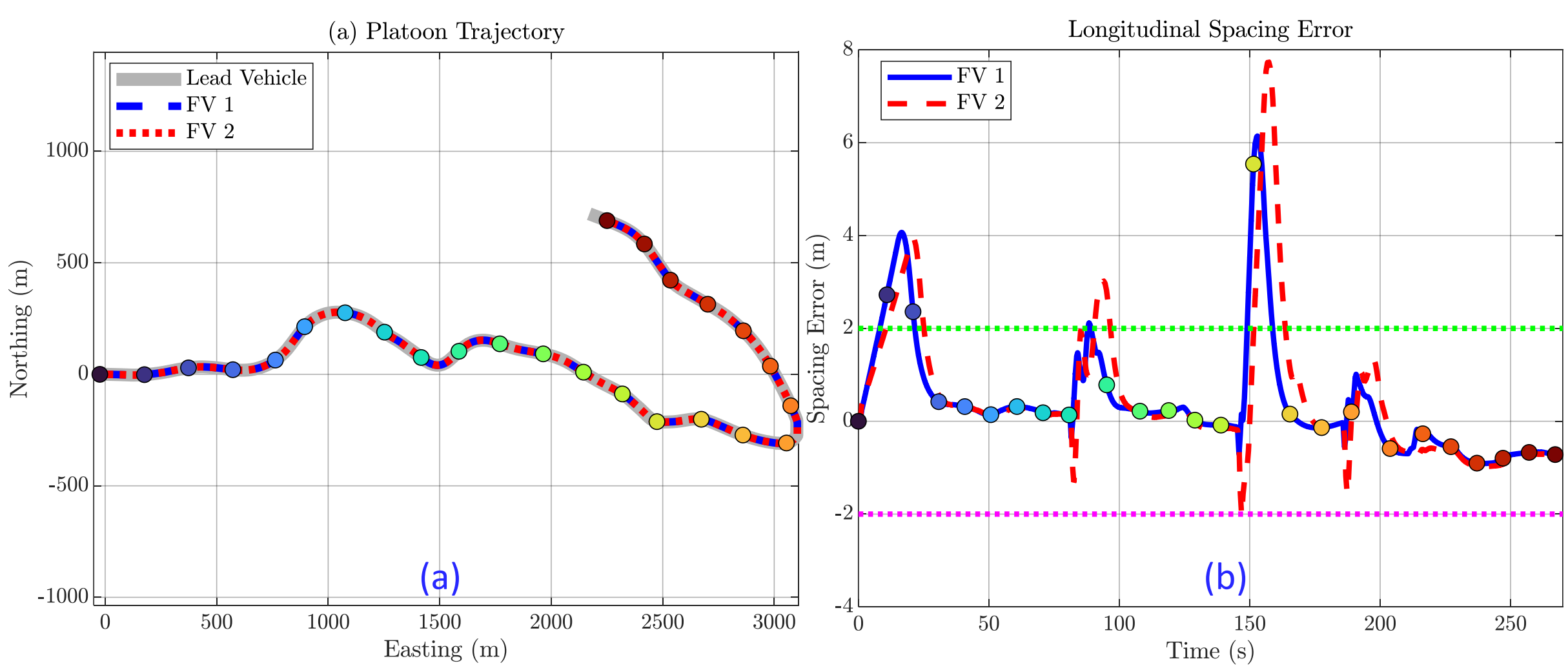

Figure 11 Longitudinal performance metrics for Blueridge Parkway platooning. (a) Platoon trajectory; (b) Spacing error showing transient deviations within ±2 m

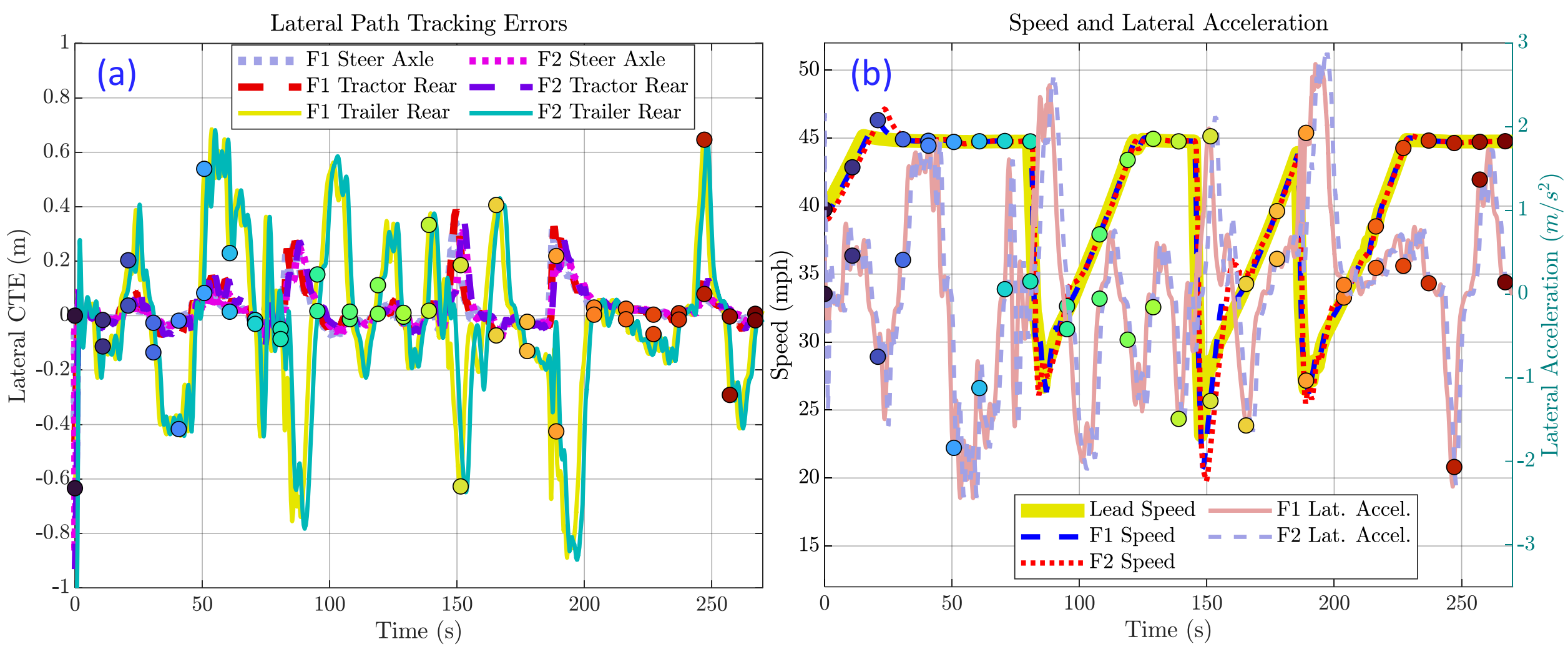

Figure 12 Lateral performance in the Blue Ridge Parkway scenario: (a) cross-track errors (CTEs) at key tractor and trailer axles, showing well-controlled errors within ±0.4 m for the tractor and larger, expected deviations up to ±0.8 m for trailer rear axles due to off-tracking; (b) dual-axis plot of follower speeds and lateral accelerations, confirming accurate speed tracking and lateral accelerations maintained within safe limits of ±0.5 g's

The lateral control remains stable and effective throughout the maneuver. As expected for articulated vehicles, the trailer rear axles exhibit the largest cross-track errors (CTEs) due to inherent off-tracking in curves, with peaks reaching approximately ±0.8 m (Figure 12a). In contrast, the CTEs for the more critical tractor steer and rear axles are well-controlled, remaining mostly within ±0.4 m. The dual-axis plot (Figure 12b) shows that the speed profile is tracked effectively by the followers, and the corresponding lateral accelerations are kept within a safe and comfortable range of ±2.5 m/s². These results confirm the integrated framework's ability to handle the combined longitudinal and lateral challenges of a realistic, high-speed, curved driving environment.

## 4.3 South Grade Road Scenario (CA)

The South Grade Road scenario evaluates the integrated framework's performance on the most demanding trajectory, featuring a series of tight hairpin turns that challenge both low-speed handling and the ability to manage extreme articulation.

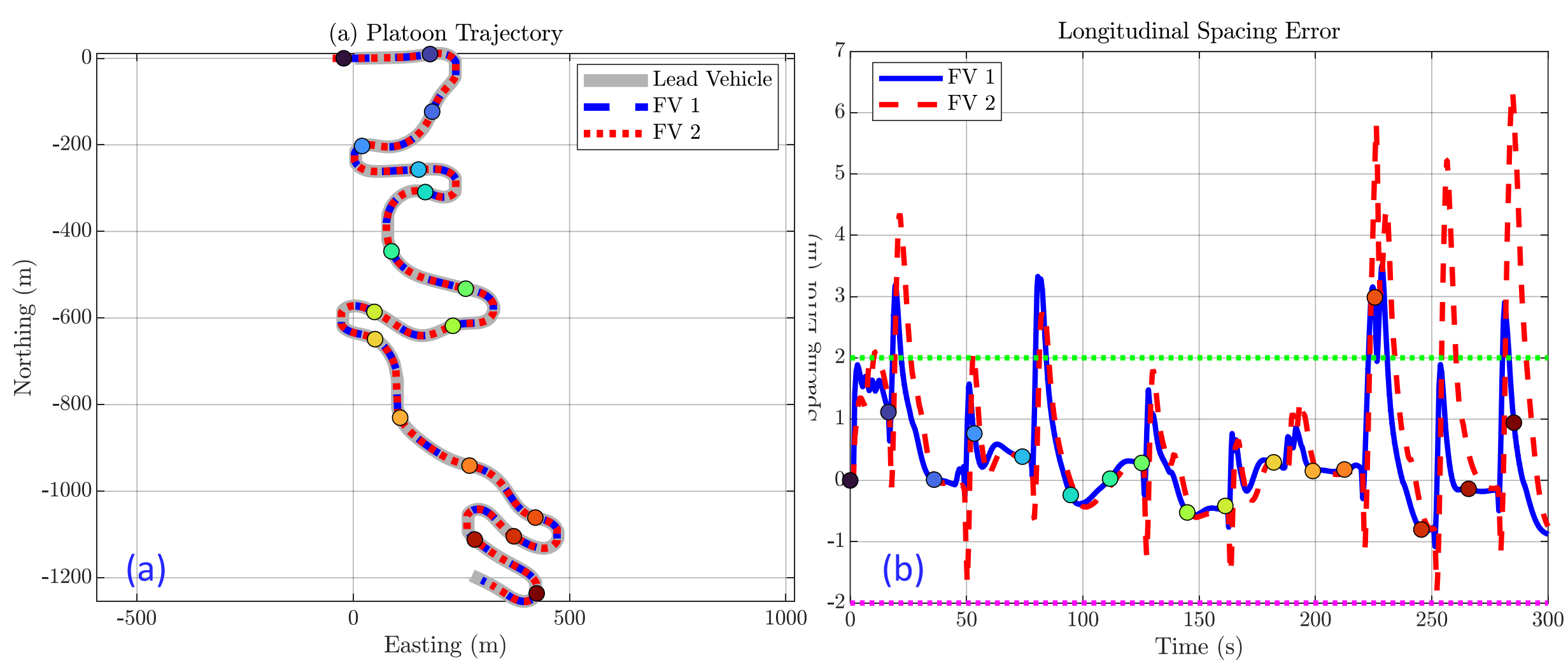

Figure 13 Longitudinal control performance on South Grade Road. (a) Platoon trajectory showing path coordinates (b) Spacing error among vehicles

As the platoon navigates the serpentine path (Figure 13a), the frequent and severe decelerations required for the turns induce significant transient spacing errors (Figure 13b). This “accordion effect” is more pronounced here than in previous scenarios due to the drastic and constant speed changes. However, the longitudinal controller ensures the platoon's safety, as the spacing error remains safely positive, and the formation recovers quickly on the intermittent straight sections.

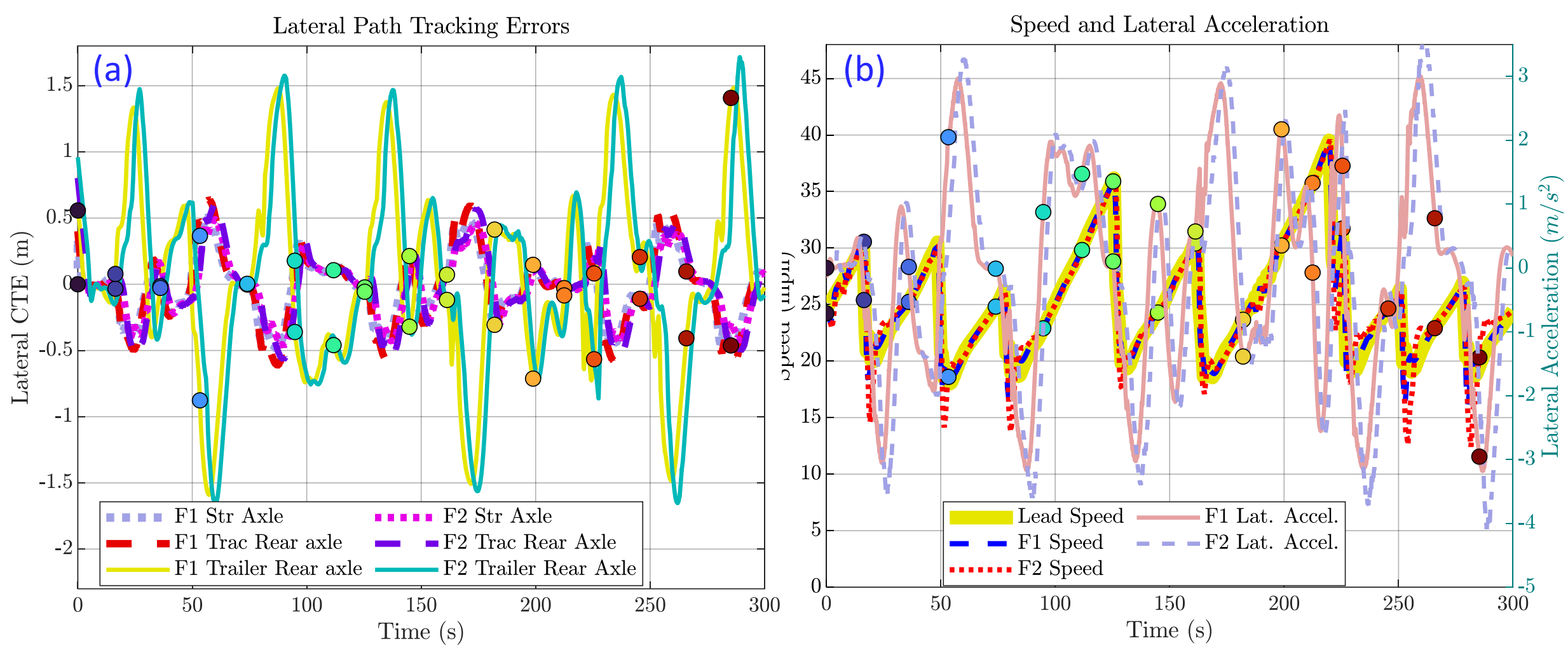

Figure 14 Lateral performance on South Grade Road: (a) Cross-track errors showing large trailer axle deviations (up to ±1.6 m) from off-tracking, while tractor errors remain within ±0.6 m. (b) Dual-axis plot of speed (15–40 mph) and lateral acceleration, confirming safe limits (± 0.5 g’s) despite frequent low-speed hairpins.

Lateral performance highlights the controller’s ability to manage extreme off-tracking, a key challenge for articulated vehicles. As expected in such tight, low-speed turns, the trailer rear axle exhibits substantial CTE, peaking at approximately ±1.6 m (Figure 14a). This represents the physical off-tracking of the trailer. In contrast, the controller effectively manages the primary vehicle unit, keeping the tractor's CTE within a much tighter bound of ±0.6 m. The speed profiles (Figure 14b) confirm the demanding nature of the task, with speeds fluctuating between 15 and 40 mph. Despite this, the corresponding lateral accelerations are consistently maintained within safe limits (±3 $m/s^2$), demonstrating the controller's stability even under the most challenging conditions.

# 5. Conclusions and Future Works

An integrated longitudinal and lateral control framework for high-speed platooning of articulated vehicles was presented, addressing challenges unique to autonomous vehicle operation in dynamic and complex scenarios. The proposed longitudinal control algorithm, based on a nested PID structure, demonstrated robust string stability through a novel proof that ensures consistent inter-vehicle spacing even under transient conditions. Additionally, the arc-length-based longitudinal error formulation expanded the algorithm's applicability to curved trajectories, enabling improved accuracy in real-world scenarios.

The lateral control algorithm, enhanced for high-speed performance, incorporated several novel adaptations, including trajectory segment classification, a dual-adaptive lookahead rate mechanism, and extensions for articulated vehicle configurations. These modifications addressed oscillations, minimized cross-track errors, and maintained stability across varying road geometries. Validation through extensive simulations demonstrated the effectiveness of the integrated framework in maintaining stability and minimizing tracking errors across a range of challenging trajectories, including high-speed maneuvers, moderate curves, and tight hairpins.

Despite these advancements, several challenges remain that warrant further investigation:

- Sensor Integration: Addressing sensor noise, inaccuracies, and GPS outages experienced in practice, using advanced filtering and fallback strategies is essential.
- Lead Vehicle Coordination: Implementing bidirectional communication to enable cooperative adaptive platooning can improve stability during complex maneuvers.
- Scalability and String Stability: Developing scalable control architectures with optimized velocity-weighting functions and incorporating time-delay effects in string stability proofs is critical.
- Field Testing: Testing across highways and urban environments will refine algorithms and prepare them for commercial deployment.

These steps will bridge the gap between simulation and practical application, enhancing the robustness and scalability of autonomous vehicle platooning systems.